\documentclass[11pt,a4paper]{article}
\usepackage[hyperref]{emnlp2020}
\usepackage{times}
\usepackage{latexsym}

\usepackage{microtype}

\aclfinalcopy 

\usepackage{graphicx}
\usepackage{subcaption}
\usepackage{amsmath}
\usepackage{mathtools}
\usepackage{amsfonts}

\usepackage{times}
\usepackage{latexsym}
\usepackage{amsmath}
\usepackage{url}
\usepackage{mathdots}
\usepackage{yhmath}
\usepackage{cancel}
\usepackage{color}
\usepackage{siunitx}
\usepackage{array}
\usepackage{multirow}
\usepackage{amssymb}
\usepackage{gensymb}
\usepackage{tabularx}
\usepackage{booktabs}
\usepackage{svg}
\usepackage{tablefootnote}
\usepackage{cleveref}
\usepackage{tabularx}
\usepackage{xcolor}
\usepackage{makecell}
\usepackage{enumitem}

\newcommand{\latentplanmethod}{\textsc{LAP}}
\newcommand{\latentplan}{\textsc{LAP-Cinf-Udec}}
\newcommand{\latentplanfull}{\textsc{LAP-Uinf-Udec}}
\newcommand{\latentplantreedecoder}{\textsc{LAP-Cinf-Cdec}}
\newcommand{\rakebaseline}{\textsc{SupervPlan}}
\newcommand{\titlebaseline}{\textsc{NoPlan-LM}}
\newcommand{\groundtruth}{\textsc{ROC-data}}
\newcommand{\cut}[1]{}

\title{Narrative Text Generation with a Latent Discrete Plan}

\author{Harsh Jhamtani $^1$ \qquad Taylor Berg-Kirkpatrick $^2$ \\
$^1$ School of Computer Science, Carnegie Mellon University\\
$^2$ Computer Science and Engineering. University of California San Diego \\
\tt{jharsh@cs.cmu.edu, tberg@ucsd.eng.edu}
}

\date{}

\begin{document}
\maketitle
\begin{abstract}
Past work on story generation has demonstrated the usefulness of conditioning on a generation plan to generate coherent stories. 
However, these approaches have used heuristics or off-the-shelf models to first tag training stories with the desired type of plan, and then train generation models in a supervised fashion. In this paper, we propose a deep latent variable model that first samples a sequence of anchor words, one per sentence in the story, as part of its generative process. 
During training, our model treats the sequence of anchor words as a latent variable and attempts to induce anchoring sequences that help guide generation in an unsupervised fashion. We conduct experiments with several types of sentence decoder distributions -- left-to-right and non-monotonic, with different degrees of restriction. Further, since we use amortized variational inference to train our model, we introduce two corresponding types of inference network for predicting the posterior on anchor words.
We conduct human evaluations which demonstrate that the stories produced by our model are rated better in comparison with baselines which do not consider story plans, and are similar or better in quality relative to baselines which use external supervision for plans. Additionally, the proposed model gets favorable scores when evaluated on perplexity, diversity, and control of  story via discrete plan.
\end{abstract}

\section{Introduction}

Maintaining long-term narrative flow and consistency are important concerns when aiming to generate a plausible story \cite{PorteousC09,hou2019survey}. Prior work on narrative text generation has focused on generating consistent stories via story outlines using keywords or key phrases \cite{ xu-etal-2018-skeleton, yao2019plan}, event-based representations \cite{RiedlY10,martin2018event, fan2019strategies}, plot graphs \cite{plotgraphs} or a sentence representing theme \cite{chen2019learning}. 

\begin{figure}[t]
    \centering
    \includegraphics[width=0.48\textwidth]{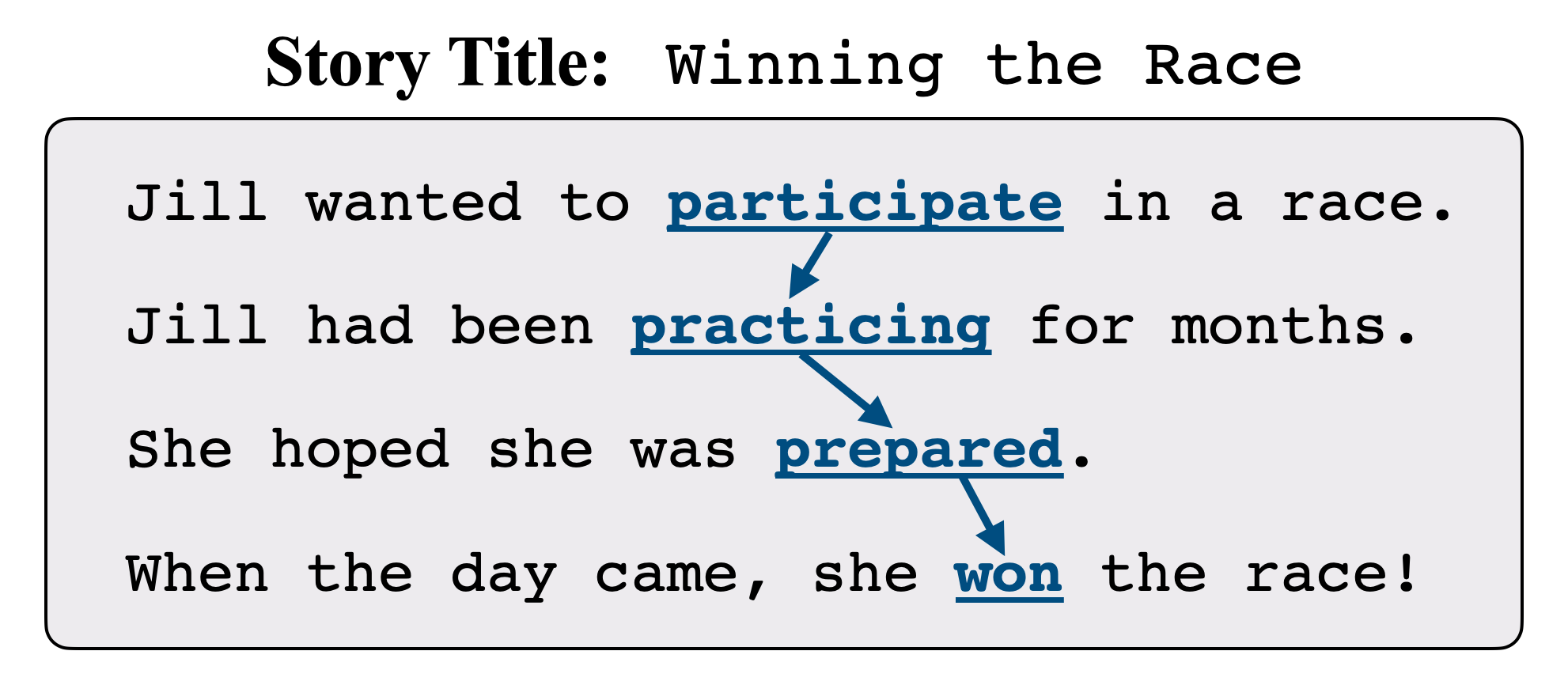}
    \caption{Our aim is to generate a story given a title. We propose models which first generate a high level story plan realized via a sequence of anchor words.}
    \label{fig:intro}
\end{figure}

\newcite{yao2019plan} note that compared to specific event based representations, using keywords to form the outline is more generalizable and widely applicable. 
In this work, we consider a sequence of anchor words as a means to model story outlines. For example, in Figure~\ref{fig:intro}, given a story title `Winning the Race', our model first predicts a sequence of anchor words which represents a high level story plan. Thereafter, a decoder conditions on the title and generated sequence of anchor words to generate the final story. We assume an alignment between the anchor words and the story sentences -- the $i^{th}$ anchor word corresponds to the $i^{th}$ sentence in the story. 

However, stories do not naturally occur with a tagged set of such anchor words or keywords. Many prior works use off the shelf tools to first label stories with plan outlines, thus using external supervision for learning plot structures. For example, \newcite{yao2019plan} use the RAKE heuristic \cite{rose2010automatic} to first identify the most important keyword in each sentence, and then use this to train a model in a supervised fashion.
This approach leads to improved coherency and control, but creates a reliance on such heuristics and does not jointly learn anchor words along with the generator. 

\begin{figure}[t]
    \centering
    \includegraphics[width=0.48\textwidth]{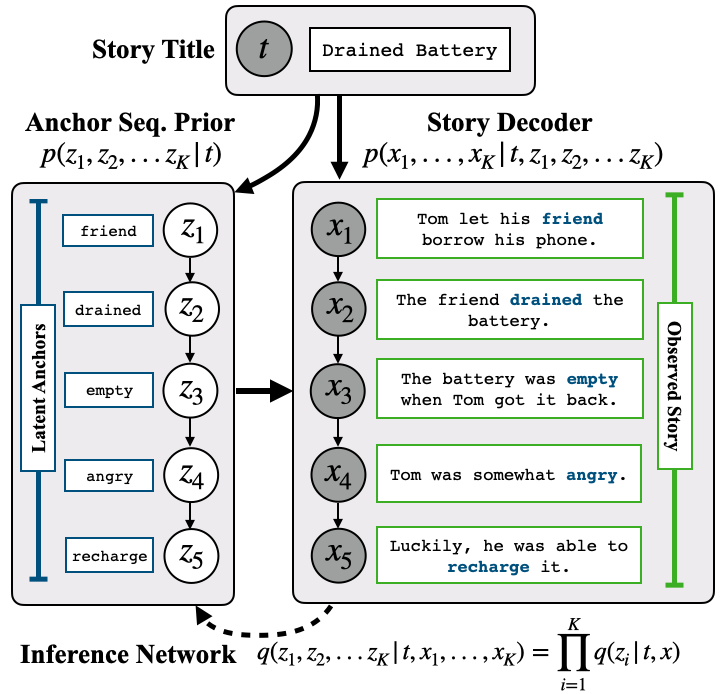}
    \caption{Model Overview: We consider multi-sentence text generation via a latent generation plan realized through a sequence of anchor words with one word per sentence. [We show sequence models with first-order Markov assumption for simplicity, even though all sequence models in our approach are auto-regressive with full context.]  }
    \label{fig:model}
\end{figure}

Inspired by prior work indicating that anchor words can effectively capture and control high-level generation structure, we investigate to what extent high-level control can be learned in a fully unsupervised fashion, directly from natural story data. We design a hierarchical latent variable model (Figure \ref{fig:model}) that induces sequences of anchor words that explain observed stories, while at the same time learning to generate entire stories by first generating anchor sequences.
For training, we use amortized variational learning \cite{kingma2013autoencoding}, where an inference network is used to approximate the posterior on anchor sequences. 

At test time, given a title, we first sample a sequence of anchor words using the prior model conditioned on only the title, and then generate the actual story using the decoder conditioning only on the title and the sampled anchor words.

\begin{figure*}[t]
    \centering
    \includegraphics[width=0.95\textwidth]{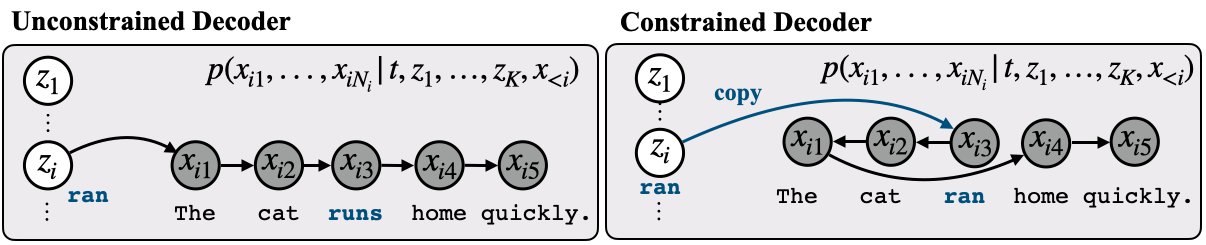}
    \caption{Simplified demonstration of generation of a sentence conditioned on anchor words and preceding sentences for the two types of decoders: (1) Unconstrained decoder is based on the story generation model of \cite{yao2019plan}, which may or may not use the corresponding anchor word. (2) Constrained decoder is forced to use anchoring words in corresponding sentences, generating words to the left and then to the right of an anchor word. 
    [Again, we show sequence models with a first-order Markov assumption for simplicity, even though all sequence models are auto-regressive with full context. ] 
    }
    \label{fig:decoders}
\end{figure*}

To induce a useful latent generation plan and to effectively condition on a sampled plan, we propose a constrained story decoder and constrained inference network. Specifically, our constrained decoder begins a story sentence by deterministic \textit{copying} the corresponding anchor word, and then generates words to the left and then to the right (Figure \ref{fig:decoders}). For this decoder, the corresponding true posterior on anchor words is sparse: the anchor word must be chosen from the observed sentence. Thus, we constrain the output vocabulary of the corresponding inference network to the words of the input sentence. 
We observe that the proposed constrained inference network does not suffer from mode collapse, leading to models which can effectively learn useful anchor words. 
Further, we also contrast this approach with a model whose decoder is not constrained to use each anchor word in each sentence. The true posterior in this case is over the full vocabulary. We conduct experiments with both constrained and unconstrained decoders and inference networks, and find that the best results are achieved through the combination of an unconstrained decoder with a constrained inference network -- indicating, perhaps, that while it is more effective to use flexible models, using a constrained inference network can add a useful inductive bias, leading the model to mimic the constraint of the inference network. 

We experiment with two English story datasets, and observe that our best models achieve favorable scores relative to several baselines when evaluated on perplexity, diversity, coherency, and controllable story generation as per various automatic and human evaluations.

Finally, we note that our modelling approach for story generation has an interesting connection with work that treats text as a latent variable in deep generative models \cite{miao2016language,wen2017latent}. We treat a latent sequence of anchor words as a form of hierarchical control over generated outputs, while related work treats the latent sequence itself as sequential text that is the output of the model.

\section{Model}

Our goal is to generate a story $x$, consisting of multiple sentences $x_1,x_2,..x_K$, given a title $t$.
Our model's generative process is depicted in Figure \ref{fig:model} and operates as follows: First, a sequence of anchor words representing a generation plan is sampled from an auto-regressive prior conditioned on the title. Next, for each anchor word, a sentence is generated conditioned on the anchor words and previously generated sentences using a decoder. During training, the sequence of anchor words is unobserved and treated as a latent variable. As described in more detail later, we will explore several choices of decoder -- those that treat anchor words as an explicit token in the sentence to be generated, generating surrounding context to the left and right, and those that simply treat the anchor words as conditioning information. In the former case, the posterior must be sparse. In the latter case, our choice of variational learning scheme will bias (but not force) the model to use anchor words in output story sentences. We shall refer to our proposed model as Latent Anchor Plan model ( \textbf{\latentplanmethod{}}). 

\subsection{Anchor Sequence Prior}
We model the sequence of anchor words representing the generation plan via a sequence of discrete random variables $z_1,z_2,..,z_K$.  
Since our aim is to induce latent plans, we assume $z$ are unobserved. We consider an auto-regressive prior model $p_\phi(z|t) = \prod_i p_\phi(z_i|z_{<i},t)$ where each anchor word is conditioned on preceding anchor words and the title $t$.

\subsection{Story Decoder}
Our decoder $p_\theta(x|t,z)$ generates a story given the title $t$ and anchor words $z$. As mentioned earlier, $z_i$ is aligned to the sentence $x_i$. We consider two decoders: (1) an unconstrained decoder which is not bound to use $z_i$ in $x_i$, and (2) a constrained decoder 
which assumes $z_i$ is present in $x_i$, and constructs words to the left and then to the right of the anchor word $z_i$.  \\

\noindent \textbf{Unconstrained Decoder:}
Our unconstrained decoder is based on \newcite{yao2019plan}'s decoder which does not use any explicit alignment of anchor words to corresponding sentences (Figure \ref{fig:decoders}).
The decoder is fed the title and the anchor words appended together, and is trained to generate the multi-sentence text.
The decoder is not bound to use the anchor word $z_i$ for $x_i$, but may have incentive to do so depending on the training objective, as discussed later. At the same time, the unconstrained decoder has higher flexibility and can skip using an anchor word if it doesn't fit with the preceding context. \\

\noindent \textbf{Constrained Decoder:}
We consider a constrained decoder that always uses $z_i$ while generating $x_i$. This is achieved by first copying $z_i$, then generating to the left until the sentence start, and then to the right. Such a decoder is bound to use the corresponding anchor word by design, and will potentially demonstrate higher control of the anchor words on the story.

Our decoder architecture follows from \newcite{yao2019plan}, who use a 3-layer LSTM recurrent model. Our final reported model uses $1000$ dimensional hidden layer, with tied input and output word embeddings. Moreover, the prior model shares the underlying LSTM modules with the decoder. 
Since our goal is to induce a latent discrete plan and compare with keyword tagging based methods, we stick to the same choice of decoder as in prior work.

\begin{figure}[t]
    \centering
    \includegraphics[width=0.40\textwidth]{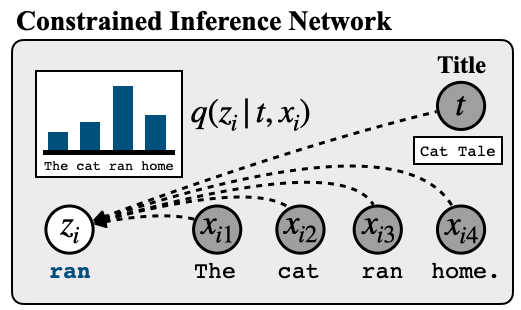}
    \caption{Constrained Inference Network: Proposed model is trained through amortized variational learning using an inference network. One of the proposed models is trained using a constrained  inference network which assigns non-zero probability to only the words present in corresponding sentences. 
    }
    \label{fig:inference}
\end{figure}

\section{Learning and Inference}
Our goal is to maximize the log likelihood of the stories conditioned on the corresponding titles. Since $z$ is unobserved at training, we must marginalize over all possible values of $z$.
\begin{equation*}
    \sum_{t,x \in \mathcal{D}} \log p(x|t) =     \sum_{t,x \in \mathcal{D}} \log \mathbb{E}_{z \sim  p_\phi(z|t)} [ p_\theta(x|t,z) ]
\end{equation*},
$\mathcal{D}$ represents the dataset of titles and corresponding stories.
Since it is infeasible to compute the exact marginal stated above, we use amortized variational learning by introducing an inference network $q_\gamma$, and train the model to maximize the following evidence lower-bound (ELBO):
\begin{equation*}
    \underbrace{\mathbb{E}_{z \sim q_\gamma(z|x,t)} [\log p_\theta(x|z,t)]}_\text{Reconstruction} - 
     \underbrace{\text{KL}(q_\gamma(z|x,t)||p_\phi(z|t))}_\text{KL-term} 
\end{equation*}
We shall refer to the first term as the reconstruction term and the second term as the KL-term.


We make a mean-field assumption in the posterior approximation on $z$ as follows: $q(z|x,t) = \prod_{i=1}^{K} q(z_i|x_i,t) $. Note that $p(z|t)$ is auto-regressive, and thus it is intractable to exactly compute the KL term. 
We resort to Monte Carlo sampling  to approximate the ELBO by drawing samples from inference network; though we will perform this differently for the KL term and the reconstruction term (more details in Section \ref{sec:optimization}). 

\subsection{Inference Network and Posterior Sparsity}

\noindent \textbf{Constrained Inference Network}
With the constrained decoder discussed earlier, the true posterior is sparse -- so making the inference net also sparse would help the  learning procedure better approximate the true posterior (Figure \ref{fig:inference}). To leverage this observation, we constrain the inference network's output distribution to have non-zero probabilities only on the tokens present in the corresponding sentence:
\begin{align*}
    q(z_i=v|x_i,t) &= 0 \text{ if } v \not \in x_i \\
    & \propto \exp(s_v) \text{ otherwise}
\end{align*}
Here, $s_v$ is the logit output for the token $v$ produced by the inference network.
Our constrained inference network is a BiLSTM model which generates an encoding $h_j$ for $j^{th}$ token in a story sentence.  A linear layer transforms $h_j$ to a score $s_j$. Finally, for sentence $x_i$, we compute a softmax over the scores of words in $x_i$ to obtain $q(z_i|x)$. \\ 

\noindent \textbf{Unconstrained Inference Network}
We also consider an unconstrained inference network which does not constrain the inference network's output -- i.e. the output distribution is over the entire vocabulary. We use a LSTM model to encode each sentence, obtain the last word hidden state, and then finally employ a linear layer to transform it to the vocabulary size.  \\ 

When the decoder is not constrained, it may be interesting to compare the choice of inference network. Using the constrained inference net with the unconstrained decoder will bias the decoder to use the anchor words in the aligned sentences -- the model is not required to do this, but variational learning will pull the inference network and true model posterior towards each other (i.e. the ELBO objective pressures them to agree). Thus, if the inference net is constrained, but the decoder is not, learning will try to find a weakly constrained decoder to match the approximate posterior.

\subsection{Optimization}
\label{sec:optimization}
\noindent \textbf{Reconstruction term:}
As mentioned earlier, we draw samples from the inference network to approximate the reconstruction term. The decoder parameters $\theta$ can be trained directly through back-propagation to minimize the approximate reconstruction loss.
However, since $z$ is discrete, we use the REINFORCE \cite{williams1992simple} algorithm to train the parameters $\gamma$ of the inference network $q(z|x,t)$.  Following prior work \cite{xu2015show}, we use an entropy regularizer term and a moving average baseline to reduce the variance of the resulting gradient estimator for inference network parameters $\gamma$. \\

\noindent \textbf{KL term:}
Note that the KL term can be simplified as follows:
\begin{align*}
    \text{KL}&(q_\gamma(z)||p_\phi(z)) = \text{KL}(q_\gamma(z_1)||p_\phi(z_1)) + \\ &\mathbb{E}_{z_1 \sim q_\gamma(z_1)} [ \text{KL}(q_\gamma(z_2)||p_\phi(z_2|z_1)) + \\ &\mathbb{E}_{z_2 \sim q_\gamma(z_2)}[ \text{KL}(q_\gamma(z_3)||p_\phi(z_3|z_{<3})] 
    + \dots  ]]]
\end{align*}

We draw samples of $z$ from $q(z)$ to approximate the KL term. \\

\noindent \textbf{KL term for the constrained inference network:}
For the constrained inference network, we have a sparse approximate posterior. Given the fact that typical sentences in our dataset are 5-20 words in length, it is computationally easy to exactly compute individual $\text{KL}(q(z_i)||p(z_i|z_{<i}))$ terms by summing over the tokens in $x_i$ instead of the whole vocabulary. This is still an approximation to the full KL term since we cannot feasibly sum over the context.
\begin{align*}
  \text{KL}(q(z_i)||p(z_i|z_{<i})  
  &= \sum_{z_i \in V}  q(z_i) \log q(z_i)/p(z_i) \\
  &= \sum_{z_i \in \bf{x_i}}  q(z_i) \log q(z_i)/p(z_i)
\end{align*}

Thus, for the constrained inference network, KL computation now proceeds as follows: we first compute $\text{KL}(q(z_1)||p(z_1))$ as described above. Then we sample $z1 \sim q(z_1)$, and compute $\text{KL}(q(z_2)||p(z_2|z_{<1}))$, and so on -- we still need to use samples, but can exactly compute each of the $K$ individual KL terms, one at each of the $K$ steps in the plan, similar to the approach of  \citep{yang2018unsupervised}. 
We observe that the constrained inference network leads to lower variance in the KL term approximation, thereby leading to more stable gradients. \\

\noindent \textbf{Pretraining:}
Pretraining the inference network in an autoencoder setup has been found useful for VAE training \cite{li-etal-2019-surprisingly}. We pretrain the inference network in an autoencoder setup where the decoder reconstructs the corresponding sentences (rather than whole story). Thereafter, we train the decoder and prior keeping the inference network fixed. Finally we perform the full training with all parameters being updated. We observe that pretraining through this procedure leads to more stable training. \\

\section{Experiments}
We evaluate and report generation quality of various models using automatic metrics for fluency and diversity, as well as human evaluations for coherence of story and relevance to title. We also analyze the ability of anchor words to control the generated story, and highlight comparisons between various choices of inference networks and decoders.

\begin{table*}[]
\small
\centering
\begin{tabular}{@{}lllccccccc@{}}
\toprule
{\bf Method} & {\bf Inference N/W} & {\bf Decoder} & \bf PPL$\downarrow$  &  \multicolumn{2}{c}{\bf NLL$\downarrow$}  & \multicolumn{2}{c}{\bf DIV$\uparrow$} & \multicolumn{1}{c}{\bf DIV-B$\downarrow$}  \\ 
& & &  \bf test & \bf test & \bf dev  & \bf plan & \bf story & \bf story  \\ 
\midrule
No Plan \\
\quad  \groundtruth{} & NA & NA &
NA & NA & NA & NA & $9.01$ & $0.23$  \\ 
\quad \titlebaseline{} & NA & Unconstrained & $\bf17.3$  & $154.0$ & $160.7$ & NA & $7.70$ & $0.50$ \\ 
\midrule
With Plan \\
\quad \rakebaseline{}  & NA \tablefootnote{We retrofit an inference network to a trained \rakebaseline{} to approximate PPL and NLL for evaluation purposes only. Training the \rakebaseline{} model does not involve any inference network.} & Unconstrained &  ${\leq}28.3$  &  ${\leq}180.3$ & ${\leq}187.6$ & $8.71$ & $7.74$ & $0.49$ \\  
\quad \latentplan{} & Constrained & Unconstrained & $\bf{\leq}21.3$  & ${\leq}168.9$ & ${\leq}176.5$ & $9.24$ & $\bf7.93$ & $\bf0.45$ \\  
\midrule
\multicolumn{3}{l}{\latentplanmethod{} other variants:} \\
\quad \latentplantreedecoder{} & Constrained & Constrained & ${\leq}\bf20.9$   & ${\leq}166.9$ & ${\leq}174.1$   & $9.24$ & $\bf7.98$ & $\bf0.44$  \\ 
\quad \latentplanfull{} & Unconstrained & Unconstrained & ${\leq}17.5$   & ${\leq}154.2$  & ${\leq}160.9$ & $0.01$ & $7.67$ & $0.52$ \\ 
\bottomrule
\end{tabular}
\caption{
Automated metrics: We report Negative Log Likelihood (NLL), perplexity (PPL) (computed using importance weighted samples for models with latent variables), and diversity (DIV and DIV-B). 
\latentplan{} performs better than \rakebaseline{} on perplexity as well as diversity. 
We also experiment with two other  variants for \latentplanmethod{}. \latentplanfull{}, which does not constrain the inference network, suffers from posterior collapse. \latentplantreedecoder{}, which uses the constrained decoder, achieves perplexity and diversity results that are comparable to \latentplan{}.  }
\label{tab:perplexity}
\end{table*}

\subsection{Dataset}
We use a subset of the ROC-stories corpus (\groundtruth{}) \cite{mostafazadeh2016corpus} used earlier by \newcite{yao2019plan}.
\newcite{yao2019plan} had chosen a subset of the original ROC corpus in order to select only those stories which are accompanied by a title.  The train, validation and test splits consist of $78529$, $9816$, and $9816$ stories respectively. Most of the data consist of five sentence stories. 
Additionally, we experiment with the visual story dataset (only the text portion), which we discuss in more detail in Section \ref{vist}.

\subsection{Methods}

\noindent \textbf{\titlebaseline}: This baseline does not consider any story generation plan and conditions only on the title. We use the same 3-layer LSTM as in the proposed model. \\

\noindent \textbf{\rakebaseline}: This baseline is based on the work of \cite{yao2019plan} which utilizes RAKE-tagged keywords as observed anchor words. The model is trained to predict the the observed anchor words and the story given the title. 
We can view this baseline as a latent variable model that was trained using RAKE keywords as the output of a deterministic inference network. 
\\ 

\noindent \textbf{\latentplanmethod}: (1) We will refer to our model with the constrained inference network and unconstrained decoder as \textbf{\latentplan}. (2) \textbf{\latentplanfull{}} uses the unconstrained inference network and unconstrained decoder.
(3) \textbf{\latentplantreedecoder{}} uses the constrained inference network with the constrained decoder. We found that the model with constrained decoder and unconstrained encoder performed poorly during training, and so we do not include it in experiments. \\

\noindent \textbf{Decoding procedure:} For all the methods, we generate samples with top-p sampling \cite{nucleus} with $p=0.6$ at the time of story generation. Unless otherwise stated, the same decoding procedure is followed for the evaluations of diversity, story quality, and controllable generation discussed below. Later in the analysis we discuss the effect of changing the parameter $p$ on some of the evaluation metrics. 

\subsection{Perplexity}
For the models with latent generation plans, we use importance weighting (IW) \cite{burda2015importance}  (with 20 samples) to estimate perplexity scores since (IW) has been shown to provide a tighter bound than ELBO for evaluation purposes \cite{li-etal-2019-surprisingly}. 
For the baseline, \rakebaseline{}, we also evaluate its marginal likelihood for comparison with our model. To do this, we separately train an inference network (with the same architecture as that used by the \latentplan{} model) to approximate the posterior on anchor words for the trained \rakebaseline{} (by keeping the trained model parameters fixed). This approximate posterior is then used to compute an upper bound on NLL and perplexity.  

The proposed model \latentplan{} performs better than the baseline \rakebaseline{}, which uses separately tagged generation plans (Table \ref{tab:perplexity}). However, the proposed method's perplexity is close to that of \titlebaseline, which does not consider any generation plan. 
This is not uncommon for deep latent variable models -- since their held-out likelihood is intractable, and most approximations yield upper bounds on perplexity, their reported perplexity is always pessimistic.
Among \latentplanmethod{} variants, we observe that \latentplanfull{} suffers from posterior collapses, and behaves similarly to  \titlebaseline{} since the latent variables $z$ are not informative or useful. 
Finally, \latentplantreedecoder{} performs similar on likelihood evaluations compared to the \latentplan{} model with an unconstrained decoder .

\subsection{Diversity}
We generate story samples for all the titles in the test split. We employ two evaluations to report diversity in the generated outputs: \\[0.4em]
\noindent \textbf{DIV} We compute the geometric mean of empirical unigram, bigram, and trigram distribution entropy from the generated set of stories \cite{jhamtani2018chess}. For methods which use generation plans, we also compute this diversity metric on anchor word sequences. Table \ref{tab:perplexity} shows the results for various models.  \latentplan{} performs better than \rakebaseline{}, achieving higher diversity for both story and plans. Among the \latentplanmethod{} variants, using the non-constrained inference network (\latentplanfull{}) leads to worse results on story diversity, and fares poorly in plan diversity (due to posterior collapse). \latentplantreedecoder{} again performs similarly to \latentplan{}.\\[0.4em]
\noindent \textbf{DIV-B} We also report inter-story BLEU4 scores \cite{selfbleu}. We compute samples from various methods for 1000 titles in the test split. For each generated story, the remaining 999 are treated as references. Thus, lower values indicate higher diversity in the generated stories.  Table \ref{tab:perplexity} shows the results.  \latentplan{} performs better than \rakebaseline{}, though is still far from the values for human written stories in the ROC dataset itself.

\subsection{Human Evaluations}
We conduct human evaluations on Amazon Mechanical Turk to evaluate the quality of generated stories given the title. We evaluate the story samples with respect to: (1) coherence, which measures the logical and coherent narrative flow in a story, and (2) fidelity to title, which measures the degree to which the story is relevant to the given title. Given two stories from two different methods, we request human annotators to provide their preference (or mark as tie).

\begin{table}[t]
\small
\centering
\begin{tabular}{@{}lll@{}}
\toprule
\bf \latentplan{} & \bf Coherence   & \bf Title-Fidelity  \\
\bf vs Method M & \bf win-tie-loss  & \bf win-tie-loss  \\ \midrule
M=\rakebaseline{}  & $0.31$ $0.37$ $0.32$ & $\bf 0.39$ $0.27$ $0.34$  \\  
M=\titlebaseline{}  & $\bf 0.36$ $0.35$ $0.29 $ $^\dagger$ & $0.33$ $0.37$ $0.30$  \\  
M=\groundtruth{} & $0.12$ $0.08$ $0.80$ $^\dagger$ & $0.08$ $0.15$ $0.77$  $^\dagger$ \\
\bottomrule
\end{tabular}
\caption{\small Human preference evaluations when pitting various methods against \latentplan{} (i.e. preference for \latentplan{} is reported under \textit{win}).
Compared to \rakebaseline{}, \latentplan{} performs better on fidelity to title and similar on coherence. 
Loss vs win judgements marked with $^\dagger$ are statistically significant under bootstrap test ($p<0.05$) considering 1000 subsets each of size 400.}
\label{tab:coherence_metrics}
\end{table}

In order to ensure the quality of human evaluations, we restrict the annotation task to annotators from Anglophone countries, and limited to workers with more than 90\% HIT (Human Intelligence Task) acceptance rates. We randomize the order of presented stories to avoid positional bias effects. Additionally, we added two `check' data points with each HIT. More specifically, to construct a `check', we pick a random story from the development set, and then prepare a `decoy' story by replacing three lines of the story with that of another randomly chosen story. The HITs where annotators marked the `decoy' as the preferred story relative to the unaltered story with respect to either coherence or fidelity for either of the two check data points are skipped. These skipped HITs are then re-annotated.

Based on the automated metrics and manual qualitative inspection, we pick \latentplan{} as the best configuration among all the variants of our model for human evaluation. 
We randomly selected 200 titles from the test split, generate samples from all the methods under consideration, and evaluate each method against \latentplan{}. Each comparison is rated by three different annotators leading to a total of 600 judgements per pair. Table \ref{tab:coherence_metrics} shows the results. 
We observe that on average, annotators found \latentplan{} outputs similar or better on coherence and fidelity compared to the baselines. 
\latentplan{} is judged better than \titlebaseline{} on coherence, perhaps because having a plan provides a rough sketch of the story leading to more coherent outputs. 
Compared to \rakebaseline{}, outputs from the proposed method \latentplan{} are judged similar in quality in terms of coherence but better in terms of fidelity to title, perhaps because the ELBO objective encourages the inference network to pick anchor words which can be more easily predicted from the title by the prior model, leading to better title fidelity.
We show example generated samples from \latentplan{} in Table \ref{table:examples}. More examples and qualitative analysis can be found in the Appendix.

We found \latentplantreedecoder{} outputs to be slightly worse than \latentplan{} and \rakebaseline{} outputs on coherency. Compared to \latentplan{}, the constrained decoder achieves slightly better scores for perplexity and diversity (Table \ref{tab:perplexity}) and control (next subsection), but suffers on overall coherency. This behavior is likely due to the reduced flexibility of the model architecture (an example output is provided in Table \ref{treedecoderexamples}). In contrast, the non-constrained decoder achieves a favorable balance between control and coherency.  This highlights an interesting balance between the generation plan and the degree to which the decoder must follow the plan.

\begin{table}[t]
\small
\centering
\begin{tabular}{@{}lcc@{}}
\toprule
\bf Method & \bf CTRL  \\ 
\rakebaseline{} & $38.8\%$ \\ 
\latentplan{}  & $\bf72.9\%$ \\ 
\midrule
\latentplanmethod{} variants: \\
\quad \latentplantreedecoder{}  &  $100.0\%$ \\ 
\quad \latentplanfull{}  & $0.0\%$ \\ 
\bottomrule
\end{tabular}
\caption{\small We evaluate models for the extent to which the story follows the generation plan by evaluating the fraction of anchor words used in corresponding sentences (CTRL). 
\latentplan{} demonstrates better control compared to \rakebaseline{}. Model with \latentplanfull{} inference network collapses, while \latentplantreedecoder{} demonstrates perfect control due to the nature of the decoder.  }
\label{tab:control}
\end{table}

\begin{table*}[]
\small
\centering
\begin{tabular}{@{}ll@{}}
    \toprule
    {\bf TITLE:} & the exam \\
    {\bf ANCHOR WORDS:} & {midterm knew nervous performed passed} \\
    {\bf STORY:} & \multirow{2}{130mm} {I had a big geometry exam today. I knew that i would have to do it. I was nervous. I had not performed since i was a little girl. I passed out. } \\
    & \\
    \midrule
    {\bf TITLE:} & the new bed \\
    {\bf ANCHOR WORDS:} & {alex new store amazing glad} \\
    {\bf STORY:} & \multirow{2}{130mm} {Alex was trying to find a new bed. She needed a new one. She went to the store to get one. She found a amazing one. She was glad she bought it.} \\
    & \\
    \midrule
    {\bf TITLE:} & picnic \\
    {\bf ANCHOR WORDS:} & {goes fancy least eating leave } \\
    {\bf STORY:} & \multirow{2}{130mm} {Last week i visited my friends to the park. It was at the fancy park. They had to eat the food and water. I had a great time eating. I had to leave.} \\
    & \\
    \bottomrule
\end{tabular}
\caption{ \small
Generated samples from the proposed method \latentplan{}. We observe that samples from the proposed method demonstrate fidelity to the title, better follow the sampled plan of anchor word sequences, and are in aggregate more coherent than baselines which do not consider a generation plan.
  \label{table:examples}}
\end{table*}

\begin{table*}[]
\small
\centering
\begin{tabular}{@{}lll@{}}
    \toprule
    \multirow{4}{20mm}{\latentplantreedecoder{}} & {\bf TITLE:} & the exam \\
    & {\bf ANCHOR WORDS:} & {failing nervous tried test shocked} \\
    & {\bf STORY:} & \multirow{2}{105mm} {Jessica was failing her math class. She was nervous to try to take the test. She tried to help. She took the test. She was shocked and confident} \\
    & & \\
    \midrule
    \multirow{4}{20mm}{\latentplanfull{}} & {\bf TITLE:} & the new bed \\
    & {\bf ANCHOR WORDS:} & {forms forms forms forms forms} \\
    & {\bf STORY:} & \multirow{2}{105mm} {Jane was about to get a new bed. She had been trying to catch a few new sheets. She decided to get a new bed. She looked at the new sheets. It was the right choice.} \\
    & & \\
    \bottomrule
\end{tabular}
\caption{ \small 
Generated samples from  \latentplantreedecoder{} and \latentplanfull{} variants of the proposed model class.
We observe that when using the constrained decoder variant, story outputs lack  coherence more often than when using the unconstrained decoder, though they demonstrate better control by design. The \latentplanfull{} variant suffers from posterior collapse, leading to a generic anchor word sequence, and often produces stories that lack overall structure.
  \label{treedecoderexamples}}
\end{table*}

\subsection{Controllable Generation}
We evaluate models for the extent to which the story follows the generation plan.
To evaluate this, we draw one story sample per title in the test split, and report the fraction of anchor words which were used in corresponding sentences  (\textbf{CTRL} ).  \latentplan{} ($73\%$) fares much better than \rakebaseline{} ($39\%$) (Table \ref{tab:control}).  
We note that in some outputs from \latentplan{}, even though the exact anchor word was not used, we observe semantically equivalent concepts being used -- for example, for the sampled anchor word `dismay', the generated story sentence was: `She then realized she wasn't able to attempt it'. 

We also analyze CTRL and DIV-B values when sampling with different values of parameter $p$ in top-p sampling. As we increase $p$, we observe higher diversity in samples, along with lower scores for CTRL for \latentplan{} as well as \rakebaseline{} (Table \ref{tab:ptrend}). This further shows an interesting trade-off between control and diversity.

\begingroup
\setlength{\tabcolsep}{3.5pt} 
\begin{table}[t]
\small
\centering
\begin{tabular}{@{}lcccccc@{}}
\toprule
p & \multicolumn{2}{c}{\latentplan{}} & \multicolumn{2}{c}{\latentplantreedecoder{}} & \multicolumn{2}{c}{\rakebaseline{}} \\
 & CTRL &  DIV-B  & CTRL &  DIV-B & CTRL &  DIV-B  \\
 \midrule
 0.5 & $80\%$  & $0.48$ & $100\%$  & $0.48$ & $43\%$  & $0.54$  \\ 
  0.6 & $73\%$ & $0.45$ & $100\%$  & $0.44$ & $39\%$  & $0.48$ \\ 
  0.7 & $67\%$ & $0.41$ & $100\%$  & $0.40$ & $34\%$  & $0.43$ \\ 
  0.8 & $59\%$ & $0.35$ & $100\%$  & $0.34$ & $29\%$  & $0.38$ \\ 
\bottomrule
\end{tabular}
\caption{Using higher $p$ in top-p sampling leads to lower control of story via plan and  higher diversity.}
\label{tab:ptrend}
\end{table}
\endgroup

\subsection{Inference Network}
The latent plan model with no constraint on the inference network, \latentplanfull, suffers from severe mode collapse and essentially ignores the plan. This demonstrates that constraining the inference network was useful in mitigating the posterior collapse issue. In preliminary experiments, we also observed that using a bag-of-words inference network 
instead of the BiLSTM leads to worse performance on perplexity, diversity and control, which indicates that the learned posteriors for the BiLSTM network are in fact considering words in context rather than just identifying topical words in the vocabulary.

On analyzing the argmax outputs from the inference network of the trained \latentplan{} model, we find that $42\%$ of the predicted anchor words are nouns, $39\%$ of them are verbs, and $11\%$ are adjectives, compared to $58\%$, $33\%$ and $6\%$ respectively for the RAKE extracted keywords for the ROC data. Thus, the inference network learned along-with the \latentplan{} model has higher preference for verbs and adjectives compared to the RAKE algorithm. \\ 


\begin{table}[]
\small
\centering
\begin{tabular}{@{}lcccc@{}}
\toprule
{\bf Model} & \multicolumn{2}{c}{\bf PPL$\downarrow$}  & \multicolumn{2}{c}{\bf DIV$\uparrow$}  \\ 
& \bf dev & \bf test & \bf plan & \bf story  \\ 
\midrule
No Plan \\
\quad  \textsc{VizStoryData} & NA & NA & NA & $8.9$ \\ 
\quad \titlebaseline{} & $\bf 38.5$ & $40.0$ & NA & $6.3$ \\ 
\midrule
With Plan \\
\quad \rakebaseline{} & ${\leq}41.5$ & ${\leq}42.2$ & $6.5$ & $6.5$ \\ 
\quad \latentplan{} & $\bf {\leq}39.9$ & $\bf {\leq}40.8$ & $8.0$ & $6.6$ \\ 
\bottomrule
\end{tabular}
\caption{
Experiments with a second story dataset. We experiment with the text portion of the Visual Story Dataset. We observe that \latentplan{} is able to perform better than \rakebaseline{} on perplexity and diversity.  }
\label{tab:vizstory}
\end{table}

\subsection{Visual Storytelling Dataset \label{vist}}
We also conduct experiments with the text portion of a visual story dataset \cite{vizstory}. The dataset consists of 40155, 4990, and 5055 stories in train, dev, and test splits. Compared to the ROC data, there are no titles associated with stories, and we learn unconditional anchor word sequence $p(z)$.
We train the best model configuration \latentplan{}  (with constrained inference network and unconstrained decoder). To train the baseline \rakebaseline{}, we run the RAKE algorithm to tag the data with the anchor words. 
We observe that \latentplan{} performs better in terms of diversity of generated stories and plans, as well as perplexity relative to \rakebaseline{} (Table \ref{tab:vizstory}). Diversity computations are performed with 200 generated samples. We provide further example generations from various methods in the Appendix.

\section{Related Work}

Prior work on story generation has largely focused on plot outline via keywords or key phrases \cite{yao2019plan, xu-etal-2018-skeleton}, event-based representations \cite{martin2018event, fan2019strategies}, or a sentence theme \cite{chen2019learning}.  \citet{liucharacter} propose a method to generate a story conditioned on a character description.
Prior work on narrative text generation with plans has mostly relied on external resources or tools to extract outlines \cite{ZhouGH18,fan2019strategies}, and then training in a supervised manner. For example, using VADER \cite{hutto2014vader} to tag sentiment polarity \cite{luo2019learning}.

Much prior work has used manually defined objectives to encourage coherence in generated text. In this context, reinforcement learning has been used to encourage stories to follow certain manually defined goals such as being locally coherent \cite{tambwekar2018controllable, xu-etal-2018-skeleton}.
Prior work on visual story generation aim to learn topically coherent visual story generation \cite{huang2019hierarchically, wang2019keep}. Compared to topics, keywords provide more fine-grained plan, and thus are more likely to provide fine-grained control over generated outputs.

In this work we have proposed a constrained inference network and a constrained decoder for story generation. Pointer networks \cite{vinyals2015pointer} have been used for amortized inference in prior work on summarization \cite{miao2016language}, though in a semi-supervised context. Non-monotonic sequence generation has been explored in past for tasks such as machine translation \cite{WelleckBDC19}.

In the proposed model, the generation plan  can be used to control the story via the anchor words. Hard and soft constraints for incorporating keywords into generation have been explored in \newcite{KiddonZC16,miao2019cgmh}. Controllable text generation has been explored in other tasks as well, such as summarization \cite{fan2018controllable}, paraphrasing \cite{goyal2020neural}, style transfer \cite{keskar2019ctrl}, and data-to-text generation \cite{shen2019select}.

\section{Conclusion}
In this work we have proposed a deep latent variable model which induces a discrete sequence of anchor words as a high-level plan to guide story generation.\footnote{\url{https://github.com/harsh19/Latent-Anchor-Plan}} We train the models though variational learning using a constrained inference network, and compare constrained and unconstrained versions of the decoder. The proposed model performs similarly or better than baselines on various automated and human evaluations.
Related approaches might be used more broadly for a variety of language generation tasks, or even for related domains like music generation. Other modeling extensions might explore richer structure in latent plans -- for example, generalizing beyond isolated words. Finally, in this work we trained decoders from scratch, though it would be interesting to explore the incorporation of large pretrained models such as GPT2 \cite{radford2019language} to increase fluency.

\paragraph{Acknowledgements}
We thank Nikita Duseja, Aashi Jain, Prakhar Gupta, Bodhisattwa P. Majumder, and the anonymous conference reviewers for providing valuable feedback. This project is funded in part by the NSF under
grants 1618044 and 1936155, and by the NEH
under grant HAA256044-17. The first author is supported in part by an Adobe Research Fellowship. Findings and observations do not necessarily reflect the views of funding agencies. 

\bibliography{emnlp2020}
\bibliographystyle{acl_natbib}

\newpage

\section*{APPENDIX}

\section*{A. Additional Implementation Details}


\noindent \textbf{Additional Training Details:}
We found it useful to add certain regularizers. Following \newcite{yao2019plan}, we add a temporal L2 penalty on successive hidden state representations of LSTM. Additionally, we block stopwords from being sampled from the posterior since we are more interested in inducing generation plans. We use NLTK's English stop-words list for this purpose. During model training (after pretraining inference network), we also use KL thresholding / free-bits \cite{aziz2020} which thresholds each component of the KL term to help prevent posterior collapse. \\

\begin{figure}[h]
    \centering
    \includegraphics[width=0.50\textwidth]{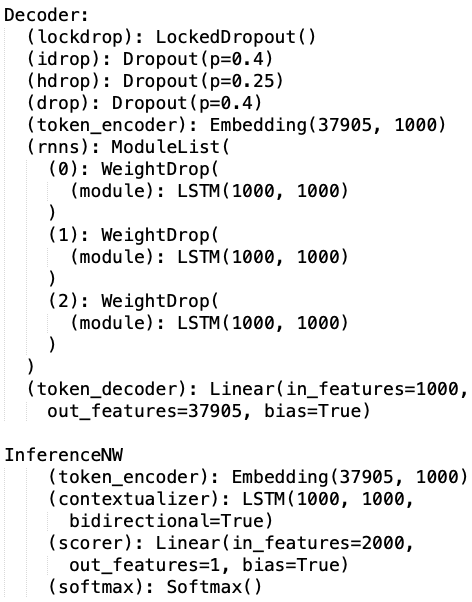}
    \caption{Summary of model architecture. }
    \label{fig:arch}
\end{figure}

\noindent \textbf{Hyperparameters} 
We perform model selection based on best dev split performance as per NLL. (In case of latent variable models, we use the upper bound on NLL). The final model and training configuration for \latentplan{} is as follows: batch size of $20$, temporal regularization weight of $1.0$, smoothing factor for moving average baseline for reinforce reward is $0.1$, dimension of hidden embedding is $1000$, input and output token embeddings are tied. A summary of the decoder and inference network for the final configuration of \latentplan{} model is shown in Figure \ref{fig:arch}. \\

\noindent \textbf{Datasets:}
We use ROC data \footnote{\url{https://cs.rochester.edu/nlp/rocstories/}} splits from \cite{yao2019plan}  \footnote{\url{https://bitbucket.org/VioletPeng/language-model/src/master/}}.
We also used Visual Story Dataset
\footnote{\url{http://visionandlanguage.net/VIST/}}


\section*{B. Generated Samples and Qualitative Analysis}
Some additional generated samples from various models are shown in Table \ref{examplesl}. 
We note that \latentplan{} plans often exhibits good control over the generated story. For example, samples S3 and S4 samples in Table \ref{examplesl} for the same title by-and-large follow the generated plan.
We do observe a certain degree of repetition in some samples e.g in sample S2, the first and third sentences both discuss mowing the lawn. 

Sample S6 further demonstrates the generation order for \latentplantreedecoder{}. Each sentence begins by copying the corresponding anchor word, generating words to the left and then to the right of it. Thus \latentplantreedecoder{} model commits to using the corresponding anchor words. 
In contrast \latentplan{} has more flexibility in using the anchor words, and sometimes anchor words get dropped or get used in a paraphrased manner. e.g. for sample S2, the model did not use the word `tired' in the last story sentence, though the second last sentence evokes similar understanding. However, this also means the control of the story via the plan is somewhat weaker. e.g. for S4, the story does not use the anchor word `boiled'. Finally, compared to outputs from \latentplan{} and \latentplantreedecoder{}, we observe that \rakebaseline{} outputs demonstrate much weaker control of the plan on the story, and generate plans less relevant to the title.

Table \ref{examplesviz} shows samples from various models trained on (text part of) visual story data. 


.

\begin{table*}[]
\small
\centering
\begin{tabular}{@{}llll@{}}
    \toprule
   \bf S.no. & \bf Method & \bf Title & \bf Story \\
   \midrule
   1 & \latentplan{} & {\bf TITLE:} & race \\
    & & {\bf ANCHOR WORDS:} & {participate high challenged sure won} \\
    & & {\bf STORY:} & \multirow{2}{100mm} {Jill wanted to  participate in a race. She was in  high school and had no idea how to race. Jill  challenged her  friend to a race. She wasn't  sure if she would win. Ultimately she won the race.} \\
    & & & \\
    & & & \\
    \midrule
    2 & \latentplan{} & {\bf TITLE:} & lawn \\
    & & {\bf ANCHOR WORDS:} & {mowing lawn mowed every tired} \\
    & & {\bf STORY:} & \multirow{2}{100mm} {I was mowing my lawn one day. I was on the lawn by myself. I mowed the lawn. I worked on it for hours. Eventually I was able to go back to work.} \\
    & & & \\
    \midrule
    3 & \latentplan{} & {\bf TITLE:} & the dryer \\
    & & {\bf ANCHOR WORDS:} & {load cold wash everything long} \\
    & & {\bf STORY:} & \multirow{3}{100mm} {A girl got a load of laundry. She put everything in the dryer. She wanted to wash it off. She did everything but the washing machine was not so good. She had to take a long time to clean it up.} \\
    & & \\
    & & \\
    \midrule
    4 & \latentplan{} & {\bf TITLE:} & the dryer \\
    & & {\bf ANCHOR WORDS:} & {cooking boiled dried put cooking} \\
    & & {\bf STORY:} & \multirow{3}{100mm} {There was a pot of beans in the microwave. The pot was piled on the table. They dried it and cooked it. They put the noodles in the oven. The kids were all cooking dinner.} \\
    & & & \\
    & & & \\
    \midrule
    5 & \latentplan{} & {\bf TITLE:} & lost phone \\
    & & {\bf ANCHOR WORDS:} & {bar opened someone left found} \\
    & & {\bf STORY:} & \multirow{2}{100mm} {Tim was at the bar. He opened the door to find a phone. It was of a person. Tim left for a couple of minutes. He took it home and it turned out that he lost it.} \\
    & \\
    \midrule 
    \midrule
    6 & \latentplantreedecoder{} & {\bf TITLE:} & race \\
    & & {\bf ANCHOR WORDS:} & {entered win losing trained win} \\
    & & {\bf STORY:} & \multirow{3}{100mm} {entered tim $<$start$>$ a local gym. win to wanted he $<$start$>$. losing up wound he $<$start$>$ the race. trained tim $<$start$>$ to get an extra. win to decided tim $<$start$>$ the race.} \\
    & & & \\
    & & & \\
    & & {\bf STORY (reordered):} & \multirow{2}{100mm} {Tim entered a local gym. He wanted to win. He wound up losing the race. Tim trained to get an extra. Tim decided to win the race.} \\
    & & & \\
    \midrule 
    \midrule
    7 & \latentplanfull{} & {\bf TITLE:} & race \\
    & & {\bf ANCHOR WORDS:} & {forms forms forms forms forms} \\
    & & {\bf STORY:} & \multirow{4}{100mm} {Cathy and his friends decided to go to the fair. Cathy was worried about the show that she was having to be on the track team. Cathy was worried that the race would have to be over. Cathy was nervous to win by a point of winning the race. Cathy caught the race.} \\
    & & & \\
    & & & \\
    & & & \\
    \midrule
    \midrule
    8 & \rakebaseline{} & {\bf TITLE:} & race \\
    & & {\bf ANCHOR WORDS:} & {today good day race race} \\
    & & {\bf STORY:} & \multirow{3}{100mm} {I was the fastest runner on my high school track team. It was very hard to win, but my friend convinced me to do it. I practiced with the runners and practiced every day. I practiced every day after school. I can go to the race every day.} \\
    & & & \\
    & & & \\
        \midrule
    9 & \rakebaseline{} & {\bf TITLE:} & lost phone \\
    & & {\bf ANCHOR WORDS:} & {package wanted found battery missed } \\
    & & {\bf STORY:} & \multirow{3}{100mm} {Tom was getting a new phone. He wanted to buy a new one. When he found it it was still a broken battery. Tom was sad because it was broken. Tom tried to save it but it wasn't enough.} \\
    & & & \\
    & & & \\
    \bottomrule
\end{tabular}
\caption{
Samples from different methods (with minor edits such as captalization) 
  \label{examplesl}}
\end{table*}

\begin{table*}[]
\small
\centering
\begin{tabular}{@{}lll@{}}
    \toprule
     \latentplan{} & {\bf TITLE:} & n/a \\
    & {\bf ANCHOR WORDS:} & {reunion together way together members} \\
    & {\bf STORY:} & \multirow{2}{110mm} {We got together to have a reunion. We went to a small town. We had a lot of fun. We took many pictures. Then we all got together to talk to some families.} \\
    & & \\
    \midrule
   \latentplan{} & {\bf TITLE:} & n/a \\
    & {\bf ANCHOR WORDS:} & {wedding wedding people food beautiful} \\
    & {\bf STORY:} & \multirow{2}{110mm} {I went to the wedding last weekend. There were a ton of people at the wedding. There were a lot of people there. The food was delicious. The cake was beautiful .} \\
    & & \\
    \midrule
    \midrule
    \titlebaseline{} & {\bf TITLE:} & n/a \\
    & {\bf ANCHOR WORDS:} & n/a \\
    & {\bf STORY:} & \multirow{2}{110mm} {This is the first day of my vacation. I like to have a trip to location. I have never been to location. Here is my apartment. I've been hoping to be there. I'm so excited.} \\
    & & \\
    \midrule 
    \midrule
    \rakebaseline{} & {\bf TITLE:} & n/a \\
    & {\bf ANCHOR WORDS:} & {big man none sat showed} \\
    & {\bf STORY:} & \multirow{2}{110mm} {The big day. And the man was playing. And the flowers were laid out. Then they sat on the floor. They were all very happy} \\
    & & \\
    \midrule
    \rakebaseline{} & {\bf TITLE:} & n/a \\
    & {\bf ANCHOR WORDS:} & {big wedding table party people} \\
    & {\bf STORY:} & \multirow{3}{110mm} {It was a big day for a wedding. The wedding party all gathered around the table. The tables were set and ready to be served. People arrived and chatted with each other. The table was set.} \\
    & & \\
    & & \\
    \bottomrule
\end{tabular}
\caption{
Samples from different methods for visual story data
  \label{examplesviz}}
\end{table*}








\end{document}